\documentclass{article}

\usepackage[preprint]{neurips_2026}

\usepackage[utf8]{inputenc} 
\usepackage[T1]{fontenc}    
\usepackage{hyperref}       
\usepackage{url}    
\usepackage{booktabs}       
\usepackage{amsfonts}       
\usepackage{nicefrac}       
\usepackage{microtype}      
\usepackage{xcolor}         
\usepackage{amsmath}
\usepackage{graphicx}
\usepackage{multicol}
\usepackage{multirow}
\usepackage{booktabs}
\usepackage{xcolor}
\usepackage{makecell}
\usepackage{makecell}
\usepackage{caption}
\usepackage[table]{xcolor}
\usepackage[export]{adjustbox}
\usepackage{pifont}
\newcommand{\Letter}{\ding{41}}
\usepackage{algorithm}
\usepackage{algpseudocode}
\usepackage{setspace}
\usepackage{placeins}
\usepackage{enumitem}
\hypersetup{hidelinks}

\title{Demo-JEPA: Joint-Embedding Predictive Architecture for One-shot Cross-Embodiment Imitation}

\author{
  \makebox[\textwidth][c]{%
    \textbf{Jingyang He}\protect\textsuperscript{\rm 1,3$^{*}$},
    \textbf{Guangrun Li}\protect\textsuperscript{\rm 1,3$^{*}$},
    \textbf{Jieyu Zhang}\protect\textsuperscript{\rm 2$^{*\dagger}$},
    \textbf{Chengkai Hou}\protect\textsuperscript{\rm 1,3},
    \textbf{Zhengping Che}\protect\textsuperscript{\rm 3},
    \textbf{Shanghang Zhang}\protect\textsuperscript{\rm 1}~\protect\textsuperscript{\Letter}
  }
  \vspace{0.2cm}\\
  \small
  \textsuperscript{\rm 1}State Key Laboratory of Multimedia Information Processing, School of Computer Science, Peking University \\
  \textsuperscript{\rm 2}University of Washington\quad
  \textsuperscript{\rm 3}Beijing Innovation Center of Humanoid Robotics
  \\
  $^{*}$ Equal contribution\quad $^{\dagger}$ Project lead\quad \protect\textsuperscript{\Letter} Corresponding author
  \vspace{0.2cm} \\
  \textbf{Project page: } \href{https://log2r.github.io/Demo-JEPA/}{https://log2r.github.io/Demo-JEPA/}
}
\begin{document}

\maketitle
\begin{figure}[!h] \includegraphics[width=\textwidth]{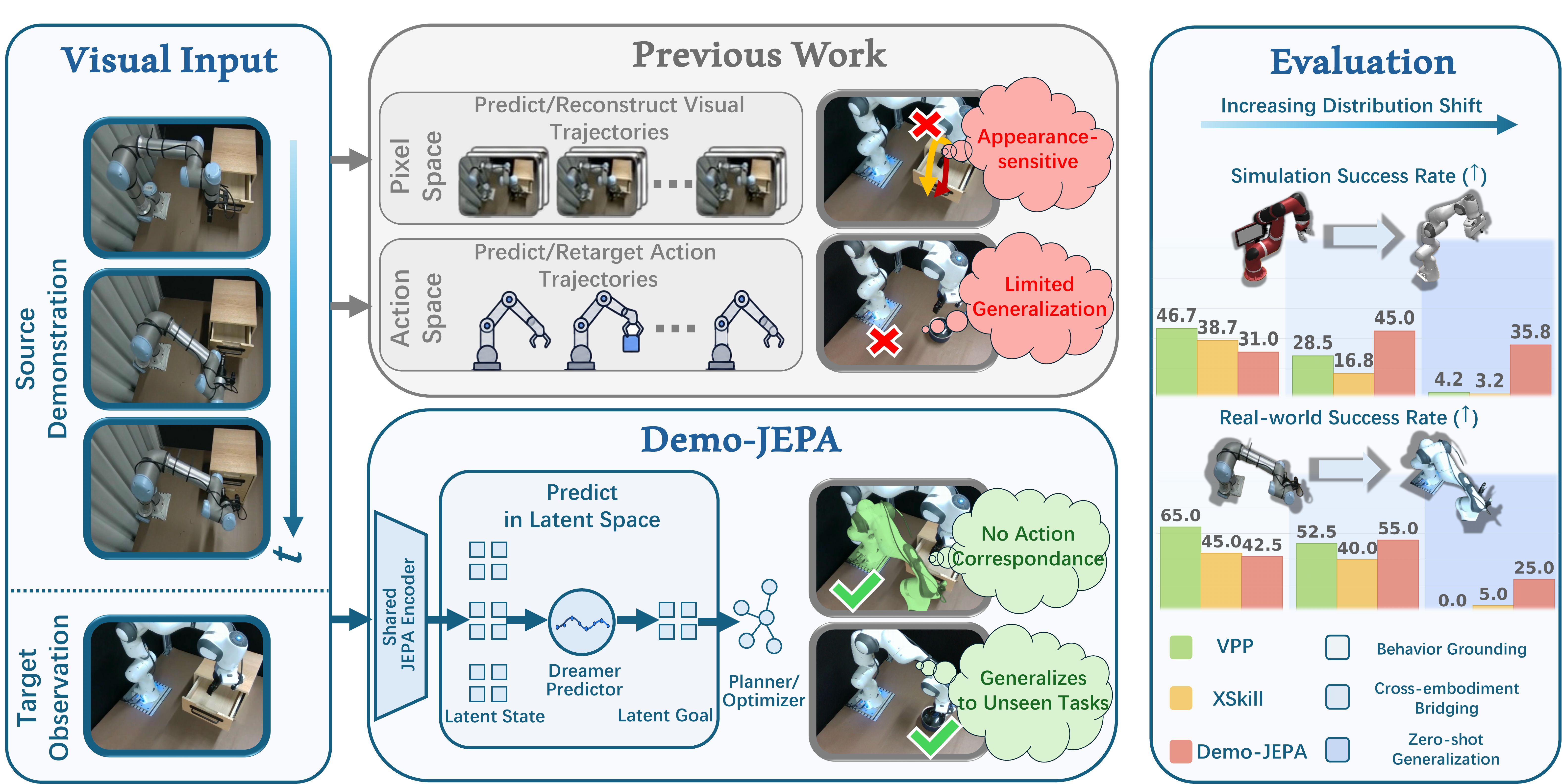} 
\caption{
\textbf{Overview of Demo-JEPA.} 
Demo-JEPA performs cross-embodiment imitation in a JEPA latent space, where the Dreamer Predictor infers target-compatible goals from source demonstrations for planning.
We evaluate it across three increasingly shifted suites: \textit{Behavior Grounding} for seen tasks, \textit{Cross-Embodiment Bridging} for unseen actions, and \textit{Zero-Shot Generalization} for unseen configurations.
Demo-JEPA achieves larger gains as distribution shift increases, showing the robustness of latent goal inference for cross-embodiment transfer.
} \label{fig:teaser} 
\vspace{-0.2cm}
\end{figure}

\begin{abstract}
Robotic imitation learning is often treated as reproducing demonstrated actions, but actions are inherently embodiment-specific.
When demonstrations come from humans or robots with different morphology, kinematics, or action spaces, this action-centric view requires shared action spaces, heuristic retargeting, or large-scale multi-embodiment co-training.
We instead view demonstrations as implicit specifications of future goals: the target agent should infer what state the demonstrator is trying to realize, rather than how the demonstrator executes it.
We propose Demo-JEPA, a cross-embodiment imitation framework that decouples demonstration intent from embodiment-specific execution.
Built on a JEPA-based world model, Demo-JEPA translates source visual demonstrations into target-compatible future latent trajectories in a shared predictive representation space.
The target agent then uses these latent trajectories as subgoals and realizes them through planning under its own learned forward dynamics.
Because Demo-JEPA avoids action-level correspondence and requires only visual demonstrations plus the target agent's own interaction experience, it supports flexible imitation across heterogeneous embodiments.
Experiments on RLBench and real-world manipulation tasks show that Demo-JEPA matches specialized in-domain planners and generalizes to unseen tasks and embodiment configurations where prior methods fail.

\end{abstract}

\vspace{-0.5cm}
\section{Introduction}
\vspace{-0.2cm}

In robotic imitation learning, a central bottleneck is the embodiment gap: the physical and kinematic mismatch between the demonstrator and the learner~\cite{torabi2018behavioral,hussein2017imitation}. Humans can often infer the intent of an observed behavior and adapt it to their own body, but robots typically struggle to translate demonstrations from another embodiment, such as a human or a robot with different degrees of freedom, into their own control space.

Many existing approaches bridge this gap through action-level alignment. Vision-Language-Action models~\cite{brohan2022rt,brohan2024rt,black2026pi0visionlanguageactionflowmodel,intelligence2025pi05visionlanguageactionmodelopenworld} and recent cross-embodiment imitation methods~\cite{zakka2021xirlcrossembodimentinversereinforcement,doshi2024scaling,fan2025xr,bauer2025latent,li2025mimicdreamer} often rely on large-scale co-training across embodiments, shared or aligned action spaces, explicit action correspondences, or heuristic retargeting. While effective in certain settings, this action-centric paradigm becomes fundamentally limited for heterogeneous agents: the same semantic behavior, such as grasping an object, may correspond to very different motor commands, torque profiles, and joint configurations across embodiments. As a result, learning cross-embodiment skills through low-level action alignment can require substantial multi-embodiment data while still struggling to generalize across mismatched morphologies.

In this work, we decouple the intent of a demonstration from its execution. Rather than asking how an action should be reproduced, we ask what future state the demonstration is trying to realize. Under this objective-centric view, a visual demonstration is not treated as a sequence of motor primitives, but as an implicit specification of a desired temporal outcome. We therefore formulate cross-embodiment imitation as latent goal-conditioned planning in a shared predictive representation space~\cite{zakka2021xirlcrossembodimentinversereinforcement,finn2017deepvisualforesightplanning,liu2022goalconditionedreinforcementlearningproblems}. Given a source visual demonstration, our framework infers embodiment-compatible future latent trajectories that serve as subgoals. The target agent then realizes these subgoals by planning under its own forward dynamics.

To instantiate this formulation, we introduce \textbf{Demo-JEPA}, a novel framework grounded in Joint Embedding Predictive Architecture (JEPA)~\cite{lecun2022path,assran2023self,bardes2024revisiting,assran2025v,murlabadia2026vjepa21unlockingdense}, as shown in Figure~\ref{fig:teaser}. 
The key design choice in Demo-JEPA is to perform cross-embodiment imitation in a predictive latent space rather than in pixel or action space. 
Unlike pixel-level generative models \cite{goodfellow2014generativeadversarialnetworks, esser2021tamingtransformershighresolutionimage, ho2020denoisingdiffusionprobabilisticmodels, song2022denoisingdiffusionimplicitmodels, lipman2023flowmatchinggenerativemodeling} or standard reconstructive autoencoders \cite{kingma2022autoencodingvariationalbayes, 10.1145/1390156.1390294, he2021maskedautoencodersscalablevision}, which may allocate substantial representational capacity to task-irrelevant details such as background textures, lighting, or embodiment-specific hardware appearance, JEPA latent spaces are optimized to capture abstract, predictive world structure. 
This property makes JEPA well-suited for cross-embodiment transfer, where the goal is to preserve high-level behavioral intent while abstracting away low-level perceptual and morphological differences.

At its core, Demo-JEPA features an embodiment-aware \textit{Dreamer Predictor} that translates visual observations from a source demonstration into target-compatible future latent states within this shared predictive space. 
These inferred latent goals then guide action generation through iterative planning in the learned latent dynamics, using the Cross-Entropy Method~\cite{rubinstein1997optimization,chua2018deep,assran2025v} (CEM). 
Crucially, Demo-JEPA does not require action-level correspondences, shared action spaces, or manually designed retargeting rules. 
The framework relies only on offline visual demonstrations from the source embodiment and the target agent's own interaction experience, reducing the need for costly cross-embodiment action annotation or data engineering.

We comprehensively evaluate Demo-JEPA in both simulated environments (RLBench~\cite{james2020rlbench}) and on real-world robotic manipulation tasks. Our empirical results demonstrate that Demo-JEPA not only matches the asymptotic performance of specialized, in-domain planners but also exhibits robust generalization capabilities to zero-shot tasks and radically different embodiment configurations where prior baselines persistently fail. By capturing the underlying predictive dynamics of a demonstration rather than memorizing surface-level statistics, Demo-JEPA successfully translates abstract task progression into an embodiment-compatible goal space, facilitating highly flexible cross-embodiment imitation.
In summary, this work makes the following principal contributions:

\begin{itemize}[leftmargin=*, itemsep=2pt, topsep=1pt, parsep=0pt, partopsep=0pt]
 \item We propose a novel formulation that reframes cross-embodiment imitation as latent goal-conditioned planning, effectively circumventing the bottleneck of explicit action alignment.
 \item We introduce Demo-JEPA, a purely vision-driven framework featuring an embodiment-aware Dreamer Predictor that bridges the morphological gap by inferring target-compatible future latent states from source demonstrations.
 \item We provide extensive validation of the framework's efficacy, demonstrating superior cross-embodiment generalization across diverse simulated and real-world robotic manipulation settings.
\end{itemize}
\section{Related Work}
\vspace{-0.2cm}
\textbf{Cross-Embodiment Imitation via Action Alignment.} 
The embodiment gap, driven by profound mismatches in morphology, kinematics, and low-level action spaces, persists as a primary bottleneck in transferring learned skills across diverse robots \cite{argall2009survey, taylor2009transfer, torabi2018behavioral, hussein2017imitation}. 
Conventional methodologies typically attempt to bridge this divide through direct action-level alignment, including explicit action retargeting and correspondence learning across embodiments \cite{gupta2017learning, devin2017learning}. 
More recent approaches construct shared action spaces or learn embodiment-agnostic control representations \cite{zheng2025universalactionsenhancedembodied, chen2026unitunifiedphysicallanguage}. 
Recent large-scale Vision-Language-Action (VLA) architectures address this challenge by leveraging massive multi-embodiment co-training datasets to implicitly internalize cross-robot control regularities \cite{brohan2022rt, brohan2024rt, embodimentcollaboration2025openxembodimentroboticlearning, kim2024openvla}. 
Other frameworks focus on discovering reusable and compositional skill primitives that generalize across embodiments \cite{xu2023xskillcrossembodimentskill}. 
Nevertheless, these action-centric paradigms face a fundamental limitation: identical semantic intents or tasks frequently demand radically different motor commands depending on the specific embodiment. 
Demo-JEPA fundamentally diverges from these approaches by entirely avoiding action-level correspondences, instead formulating cross-embodiment imitation strictly as semantic intent alignment within a shared, abstract latent space.

\textbf{Imitation from Observation.} 
Learning from visual-only demonstrations aims to recover executable control policies without access to expert action labels \cite{torabi2018behavioral, burnwal2025learningobservationsurveyrecent, chen2024miragecrossembodimentzeroshotpolicy, xie2025latentdiffusionplanningimitation, kedia2025oneshotimitationmismatchedexecution}. 
A dominant paradigm is predictive modeling, where future observations serve as a surrogate supervisory signal. 
Methods such as Visual Foresight and Video Prediction Policy (VPP) \cite{finn2017deepvisualforesightplanning, hu2025videopredictionpolicygeneralist}, as well as latent world models \cite{hafner2020dreamcontrollearningbehaviors}, learn action-conditioned dynamics to generate future trajectories and derive control by optimizing toward desired visual outcomes. 
However, these approaches typically operate in pixel space, making them computationally expensive and prone to compounding errors over long horizons. 
More importantly, pixel-level prediction often captures irrelevant visual details rather than task-relevant structure, limiting generalization under embodiment variations. 
Demo-JEPA instead performs goal inference in the latent space of a Joint Embedding Predictive Architecture (JEPA) \cite{lecun2022path}, suppressing nuisance visual variation and abstracting away embodiment-specific noise.

\textbf{World Models and Latent Goal Planning.} 
World models enable agents to simulate environment dynamics and plan in learned latent spaces, significantly improving efficiency and generalization \cite{ha2018world, ye2026worldactionmodelszeroshot}. 
Traditional approaches often rely on pixel reconstruction objectives, which can hinder abstraction and scalability. 
In contrast, JEPA-based methods focus on predicting latent representations instead of reconstructing pixels, leading to more semantically meaningful and robust representations for downstream reasoning \cite{assran2023self, assran2025v, murlabadia2026vjepa21unlockingdense}. 
Our framework builds upon V-JEPA 2.1 \cite{murlabadia2026vjepa21unlockingdense}, which learns structured, action-conditioned latent dynamics from videos. 
Within this paradigm, we integrate latent representation learning with goal-conditioned planning using sampling-based optimization such as the Cross-Entropy Method (CEM) \cite{rubinstein1997optimization}. 
By inferring target-compatible latent goals from visual demonstrations, Demo-JEPA enables the robot to execute behaviors using its own dynamics model, effectively bridging the gap between high-level intent extraction and embodiment-specific control.

\begin{figure*}[!t]
    \centering
    \vspace{-1.0cm}
    \includegraphics[width=\textwidth]{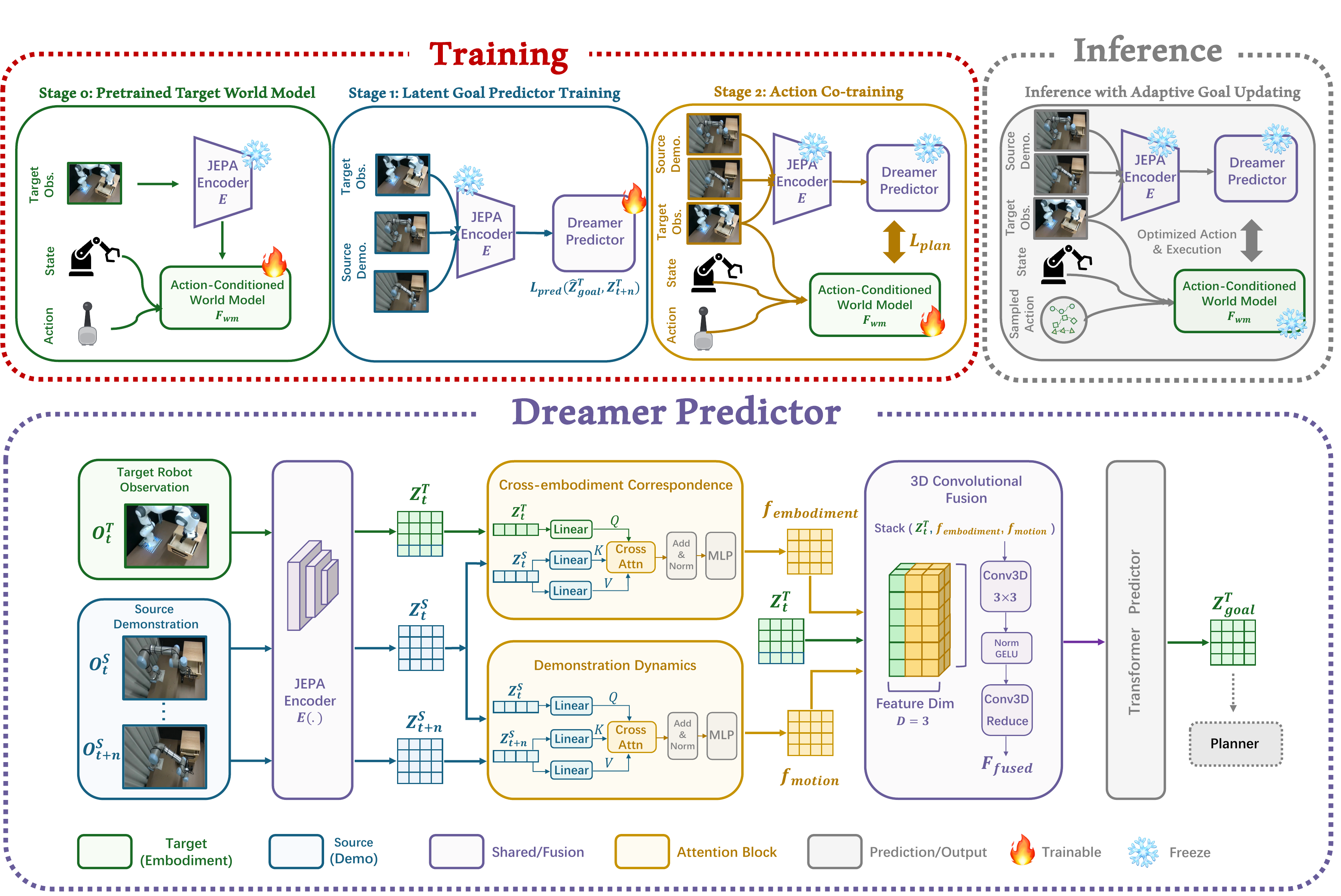}
    \caption{
    \textbf{Demo-JEPA training and inference pipeline.} The top panels show the overall training and inference stages, from target world-model initialization to closed-loop planning with adaptive goal updating.
The bottom panel highlights the Dreamer Predictor, which uses JEPA latents, cross-attention, and 3D convolutional fusion to translate source demonstrations into target-compatible future latent goals.
    }
    \vspace{-0.5cm}
    \label{fig:dreamer_predictor}
\end{figure*}
\section{Method}
\vspace{-0.2cm}
 We propose Demo-JEPA, a framework that formulates cross-embodiment imitation as latent goal-conditioned planning in a learned world model. We first define the problem setting, then introduce the latent dynamics predictor for embodiment-aware goal inference, and finally present the implementation details, including training objectives and the inference procedure.
 
\subsection{Problem Formulation}

We consider a cross-embodiment imitation setting with a source embodiment $\mathcal{E}^s$ and a target embodiment $\mathcal{E}^t$. The source provides a demonstration trajectory $\tau^s = {\{o_k^s\}}_{k=1}^{T}$ consisting only of observations. The target embodiment $\tau^t = {\{o_k^t, s_k^t, a_k^t\}}_{k=1}^{T}$ interacts with the environment through observations $o_k^t$, states $s_k^t$, and actions $a_k^t \in \mathcal{A}^t$, where the two embodiments may differ in morphology, kinematics, action spaces, and visual appearance.

At each time step $k$, Demo-JEPA takes as input the current target observation $o_k^t$ and a pair of source frames $(o_k^s, o_{k+n}^s)$ sampled from $\tau^s$, where $n$ denotes a temporal offset aligned with the planning frequency. The model outputs a sequence of target actions $a_{k:k+H-1}^t$. Instead of directly predicting actions from demonstrations, Demo-JEPA first infers a target-compatible latent goal $z_{\text{goal}}^t$ from the source frame pair and the current target observation, representing the intended future state implied by the demonstration. An action-conditioned world model then generates the action sequence by optimizing, via the Cross-Entropy Method, action rollouts whose predicted latent states match $z_{\text{goal}}^t$.

\subsection{Demo-JEPA}

Demo-JEPA formulates cross-embodiment imitation as latent goal-conditioned planning within a learned world model with a three-stage training procedure. Figure~\ref{fig:dreamer_predictor} presents the overall framework: given a source demonstration and the current target observation, the Dreamer Predictor infers an embodiment-compatible latent goal, which is then achieved through latent planning in action-conditioned world model. We now describe action-conditioned world model and dreamer predictor in detail.

 \textbf{Action-conditioned World Model.} 
We adopt VJEPA2.1~\cite{murlabadia2026vjepa21unlockingdense} as the action-conditioned world model to capture the dynamics of the target embodiment in a latent space. An JEPA encoder $E(\cdot)$ maps observations to latent states $z_k = E(o_k^t)$, and a dynamics predictor $F_{wm}(\cdot)$ models transitions conditioned on robot states and actions:
\[
\hat{z}_{k+1} = F_{wm}(z_k, s_k^t, a_k^t).
\]
Given a latent goal $z_{\text{goal}}^t$, actions are obtained by optimizing a sequence $\mathbf{a}_{k:k+H-1}^t$ such that the predicted latent rollout matches the goal:
\[
\mathbf{a}_{k:k+H-1}^{t*}
=
\arg\min_{\mathbf{a}}
d\big(F_{wm}(z_k, s_k^t, \mathbf{a}),\; z_{\text{goal}}^t\big),
\]
where $d(\cdot,\cdot)$ denotes a latent distance metric. We solve this optimization using the Cross-Entropy Method (CEM). See Appendix~\ref{app:cem} for details and Algorithm~\ref{alg:cem} for pseudocode.

\textbf{Dreamer Predictor.}
Directly imitating source trajectories is fundamentally challenging due to differences in morphology, kinematics, and interaction dynamics across embodiments. Instead, we aim to infer a future latent state that is compatible with the target embodiment while preserving the semantic intention of the source demonstration.
We introduce the \textbf{Dreamer Predictor}, an embodiment-aware latent predictor that performs cross-embodiment future-state translation in a shared JEPA representation space.
At each time step $k$, the predictor takes as input the current target observation $o_k^t$ and a source frame pair $(o_k^s, o_{k+n}^s)$, where $n>0$ denotes a temporal offset capturing future motion in the demonstration. A shared encoder maps observations into latent representations $(z_k^t, z_k^s, z_{k+n}^s)$.

The predictor jointly models two factors: (1) cross-embodiment correspondence between source and target, and (2) temporal evolution of the demonstrated behavior. We capture these via two cross-attention modules:
\[
f_{\text{emb}} = \mathrm{Attn}(Q = z_k^t,\; K = z_k^s,\; V = z_k^s), \quad
f_{\text{mot}} = \mathrm{Attn}(Q = z_{k+n}^s,\; K = z_k^s,\; V = z_k^s).
\]

The resulting features preserve the spatial structure of the JEPA latent space. We obtain the fused feature representation $f_{fused}$ by aggregating $z_k^t$,$f_{\text{emb}}$ and $f_{\text{mot}}$ through a 3D convolution $\phi$:
\[
    f_{\text{fused}} = \phi \left( \left[ \mathbf{z}_k^t \oplus f_{\text{emb}} \oplus f_{\text{mot}} \right] \right)
\]

A Transforme predictor $\mathcal{T}$ then decodes the fused feature into latent goal $
\hat{z}_{\text{goal}}^t =
\mathcal{T}(f_{\text{fused}}).
$
This latent goal serves as the objective for downstream planning in the action-conditioned world model.

\subsection{Implemention Details}

 We now describe the implementation details of Demo-JEPA, including training procedure and inference pipeline.
 \textbf{Stage 0} first pretrains an action-conditioned world model (not the focus of this work; details in Appendix~\ref{app:wm}). We focus on \textbf{Stage 1} and \textbf{Stage 2}, as well as the inference pipeline.

\textbf{Stage 1: Latent Goal Predictor Training.}
In the first stage, we train the Dreamer Predictor using action-free cross-embodiment demonstrations. The objective is to map source-embodiment transitions into target-compatible latent goals within the shared JEPA representation space. Given a target future representation $z_{k+n}^t$, the predictor is optimized via a latent reconstruction loss:
\[
\mathcal{L}_{\text{pred}} = \left\| \hat z_{\text{goal}}^t - z_{k+n}^t \right\|_2^2.
\]

To enhance generalization against execution errors and imperfect planning trajectories during inference, we introduce \textbf{temporal perturbation}. Instead of sampling a fixed source frame pair $(o_k^s, o_{k+n}^s)$, we perturb the current timestamp to obtain a randomly shifted pair $(o_{k+\delta}^s, o_{k+n}^s)$, where $\delta \sim \mathcal{U}(-r, r)$ with $r$ being the perturbation radius. This temporal regularization exposes the predictor to diverse, off-distribution latent transitions, forcing it to maintain robust goal alignments. During this stage, only the Dreamer Predictor is optimized.

\textbf{Stage 2: Action Co-Training.}
Once the Dreamer Predictor converges, it is frozen to provide stable latent goal guidance. We then unfreeze the dynamics predictor module $F_{wm}$ of the action-conditioned world model. The objective is to adapt the world model's latent rollouts to the goal distribution of Dreamer Predictor outputs. 
Given inferred goal $\hat z_{\text{goal}}^t$, the world model predicts future latent goals conditioned on sampled action sequences $\mathbf{a}_{k:k+n-1}^t$. We optimize $F_{wm}$ using a planning loss:
\[
\mathcal{L}_{\text{plan}} = \left\| F_{wm}(z_k^t, s_k^t, \mathbf{a}_{k:k+n-1}^t) - \hat z_{\text{goal}}^t \right\|_2^2.
\]
This co-training stage closely aligns the action-conditioned dynamics with the cross-embodiment intentions, significantly improving downstream generation stability. See Algorithm~\ref{alg:train} for pseudocode of training pipeline.

\textbf{Inference with Adaptive Goal Updating.}
During inference, the source demonstration $\tau^s$ serves as a sequence of reference segments. At each interaction step, the Dreamer Predictor infers a latent goal $\hat z_{\text{goal}}^t$ based on the current target observation $o_k^t$ and a source reference pair $(o_i^s, o_{i+\Delta}^s)$. Target actions are then generated via CEM optimization in the world model. Directly advancing the source reference segment at a fixed frequency often leads to compounding errors due to cross-embodiment kinematic differences. Therefore, we introduce an \textbf{adaptive goal-updating mechanism}. After executing the planned actions, we compute the latent discrepancy between the newly observed target state $z_{k+1}^t$ and the previously intended goal $\hat z_{\text{goal}}^t$ as $D_k = d(z_{k+1}^t, \hat z_{\text{goal}}^t)$.
The system advances to the next source reference pair $(o_{i+1}^s, o_{i+1+\Delta}^s)$ only if $D_k < \epsilon$, where $\epsilon$ is a predefined distance threshold indicating the sub-goal has been reached. Otherwise, the current goal remains active ($\hat z_{\text{goal}}^{t} \leftarrow \hat z_{\text{goal}}^{t}$), the Dreamer Predictor is bypassed for the current step, and the planner continues optimizing actions toward the unreached goal. This mechanism effectively stabilizes long-horizon tasks by preventing premature temporal advancement. Detail inference pseudocdeo can be found in Algorithm~\ref{alg:inference}.

\section{Experiments}\label{sec:exp}
\vspace{-0.2cm}
We validate Demo-JEPA through cross-embodiment transfer experiments in simulation and real-world environments. Our evaluation assesses its ability to infer executable latent goals from heterogeneous demonstrations without action-level correspondence, and tests its generalization across embodiments, tasks, and domains.

\begin{figure*}[t]
    \centering
    \vspace{-1.0cm}
    \includegraphics[width=\textwidth]{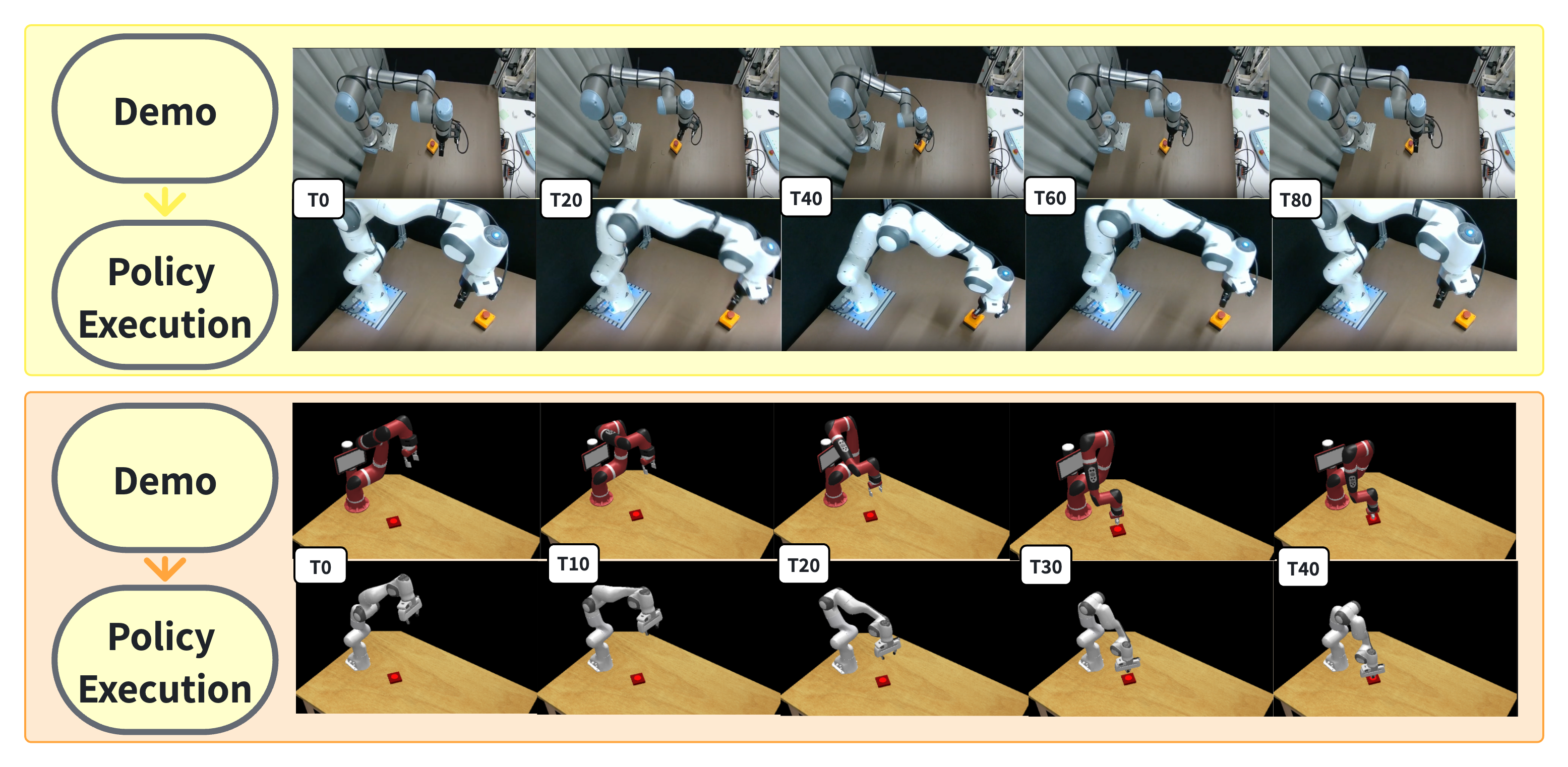}
    \caption{
    \textbf{From demo to policy execution.} Policy takes demo as input and executes action step by step. Top: real world deployment. Bottom: simulation deployment.
    }
    \vspace{-0.4cm}
    \label{fig:time_execution}
\end{figure*}
\begin{table}[!t]
\centering
\caption{\textbf{Pretraining data across transfer settings.} Table summarizes the training objective and data scale used in each stage. 
For each setting, all compared methods use the same number of tasks and trajectories, while differing in their Stage I and Stage II training objectives.}
\label{tab:pretraining_data}
\resizebox{\linewidth}{!}{
\begin{tabular}{llcccccc}
\toprule
\multirow{2}{*}{Setting} 
& \multirow{2}{*}{Method} 
& \multicolumn{3}{c}{Stage I} 
& \multicolumn{3}{c}{Stage II} \\
\cmidrule(lr){3-5} \cmidrule(lr){6-8}
& 
& Training stage & Tasks & Trajectories 
& Training stage & Tasks & Trajectories \\
\midrule

\multirow{3}{*}{Simulation}
& Demo-JEPA 
& Dreamer Predictor 
& \multirow{3}{*}{86} 
& \multirow{3}{*}{13444} 
& Action Co-Training 
& \multirow{3}{*}{39} 
& \multirow{3}{*}{8324} \\

& VPP 
& Predictive visual representation learning 
&  &  
& Action learning 
&  &  \\

& XSkill 
& Skill discovery 
&  &  
& Skill transfer 
&  &  \\

\midrule

\multirow{3}{*}{Real-world}
& Demo-JEPA 
& Dreamer Predictor 
& \multirow{3}{*}{22} 
& \multirow{3}{*}{4508} 
& Action Co-Training 
& \multirow{3}{*}{19} 
& \multirow{3}{*}{3903} \\

& VPP 
& Predictive visual representation learning 
&  &  
& Action learning 
&  &  \\

& XSkill 
& Skill discovery 
&  &  
& Skill transfer 
&  &  \\





\bottomrule
\end{tabular}}
\end{table}

\textbf{Expert Demonstration.}
In simulation, we use RLBench~\cite{james2020rlbench} to collect cross-embodiment demonstrations, acquiring video prompts from a Sawyer arm (source) to guide and evaluate a Franka arm (target) policy. For real-world experiments, we collect visual prompts using a UR5e arm to guide a physical Franka arm. As shown in Figure ~\ref{fig:time_execution}, the target policy takes the source demonstration as input and executes actions in both simulation and real-world deployments. This setup allows us to assess cross-embodiment transfer capabilities in both controlled and practical reality-gap scenarios. 

\begin{figure*}[t]
    \centering
    \includegraphics[width=\textwidth]{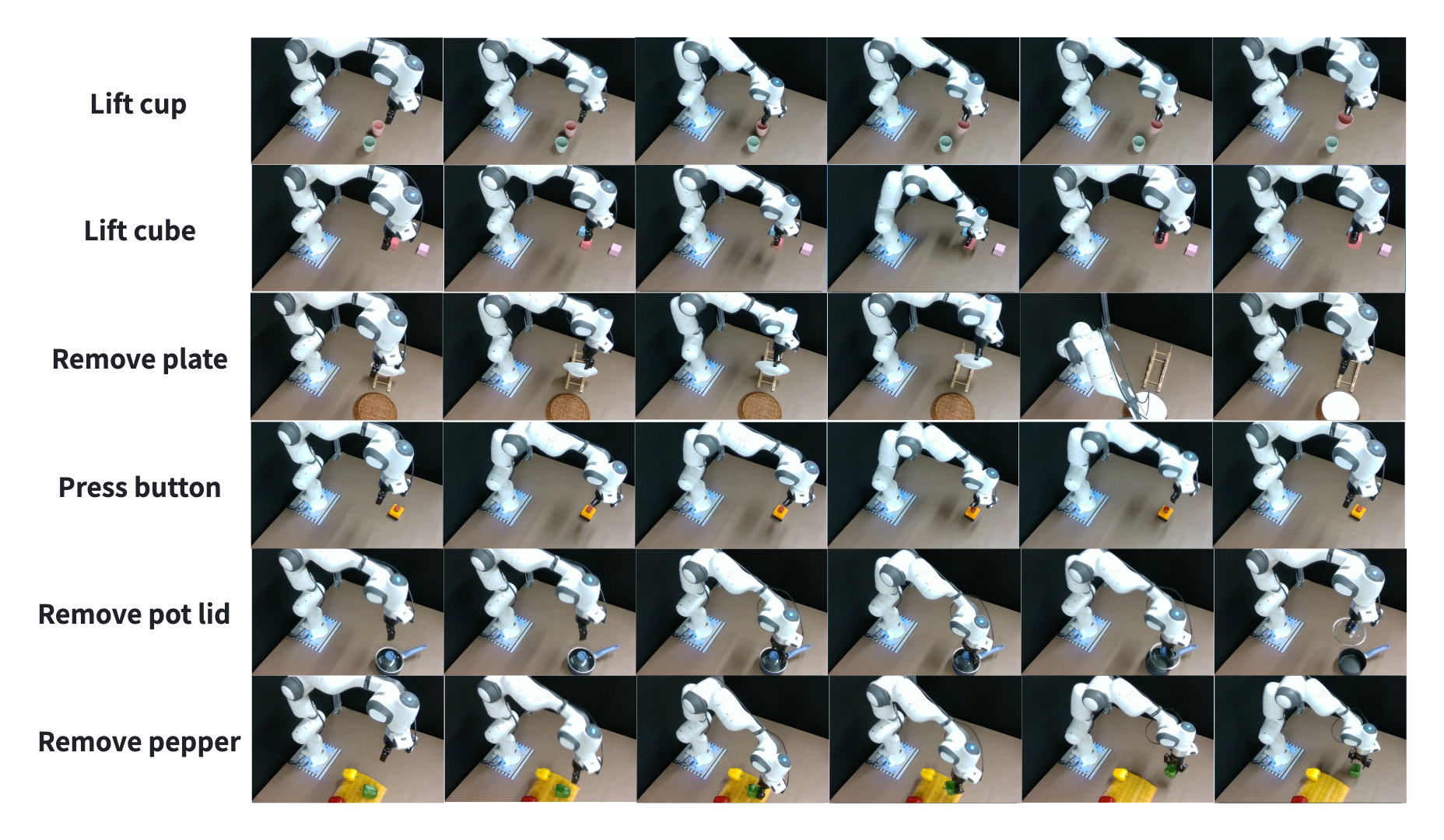}
    \caption{
    \textbf{Real-world Tasks.} We visualizes the progression of six real-world manipulation tasks.
    }
    \label{fig:task_progress}
\end{figure*}

\textbf{Task Definitions.} To evaluate diverse facets of robotic perception and control, we define six representative real-world manipulation tasks: 
(i) \textit{Lift cup} requires a stable rim-grasp on a pink cup followed by vertical elevation; 
(ii) \textit{Lift cube} involves the precise picking of a red cube within the operational space; 
(iii) \textit{Remove plate} entails grasping a plate and relocating it onto a bamboo mat; 
(iv) \textit{Press button} necessitates the targeted actuation of a small tactile button; 
(v) \textit{Remove pot lid} involves lifting and clearing a lid from its container; and 
(vi) \textit{Remove pepper} requires grasping and displacing a pepper to the workspace's lateral boundary. 
Collectively, these tasks encompass a wide array of object morphologies and contact dynamics, demanding high-precision grasping, stable manipulation, and robust spatial reasoning. Task progressions are visualized in Figure ~\ref{fig:task_progress}.

\textbf{Comparative Baselines.}
We compare Demo-JEPA with two cross-embodiment imitation baselines: VPP~\cite{hu2025videopredictionpolicygeneralist}, which conditions policy learning on future-aware visual representations from a fine-tuned video prediction model, and XSkill~\cite{xu2023xskillcrossembodimentskill}, which executes skill sequences inferred by matching heterogeneous videos to shared skill prototypes.

\textbf{Training Detail.}
Action-conditioned training uses Franka execution trajectories (camera observations, robot states, and actions) collected via the RLBench motion planner in simulation and teleoperation in the real world. 
Dreamer Predictor training utilizes paired visual trajectories (camera observations only) between the source and target (Franka) embodiments. In simulation, temporally matched Sawyer and Franka pairs are generated via retargeted replay of end-effector pose trajectories. For real-world data, where UR5e and Franka demonstrations are independent, we extract progress-aware features using GTCC~\cite{donahue2024learning} and align them in the feature space for frame-level correspondence.
The predictor is trained on these aligned pairs, leaving held-out prompts for evaluation where the inferred actions are executed by the Franka policy. Table~\ref{tab:pretraining_data} summarizes the two-stage training protocol: Stage I covers representation/transfer pretraining (Dreamer Predictor for Demo-JEPA, predictive visual learning for VPP, and skill discovery for XSkill), and Stage II handles target-embodiment action learning (action co-training  for Demo-JEPA, action learning for VPP, and skill transfer for XSkill).

\textbf{Evaluation Protocol.}
We evaluate the target Franka policy on held-out source prompt videos, reporting task success rates over \textbf{30} simulation and  \textbf{20} real-world rollouts per scenario under varied initial conditions. Based on the data protocol (Table~\ref{tab:pretraining_data}), evaluations span three suites with increasing distribution shifts: \textbf{1. Behavior grounding:} Evaluates scenarios from both stages to measure execution under familiar, fully supervised conditions. \textbf{2. Cross-embodiment bridging:} Uses Stage I scenarios to test if learned predictions support execution without direct action supervision. \textbf{3. Zero-shot generalization:} Assesses the system's ability to follow novel demonstrations under entirely unseen configurations.

\subsection{Main Results}
\begin{table}[!t]
\centering
\caption{\textbf{Simulation evaluation results.} 
We report task success rates across three simulation evaluation suites.
The best result for each task or suite average is highlighted in blue, and subsequent result tables follow the same convention.}
\label{tab:eval_results}
\resizebox{\linewidth}{!}{
\begin{tabular}{lcccccccccccccc}
\toprule
\multirow{2}{*}[-1.8ex]{\makecell[c]{\textbf{Method}}}
& \multicolumn{4}{c}{\textbf{Behavior Grounding}}
& \multicolumn{5}{c}{\textbf{Cross-Embodiment Bridging}}
& \multicolumn{5}{c}{\textbf{Zero-Shot Generalization}} \\
\cmidrule(lr){2-5} \cmidrule(lr){6-10} \cmidrule(lr){11-15}
& \makecell{Basketball\\in Hoop}
& \makecell{Change\\Channel}
& \makecell{Close\\Box}
& \makecell{\textbf{Avg.}}
& \makecell{Push\\Button}
& \makecell{Pick\\and Lift}
& \makecell{Slide Block\\to Target}
& \makecell{Pick up\\Cup}
& \makecell{\textbf{Avg.}}
& \makecell{Phone\\on Base}
& \makecell{Close Laptop\\Lid}
& \makecell{Put Rubbish\\in Bin}
& \makecell{Close\\Drawer}
& \makecell{\textbf{Avg.}} \\
\midrule
VPP~\cite{hu2025videopredictionpolicygeneralist}
& 0.53
& \colorbox[HTML]{E0F4FF}{\textbf{0.27}}
& \colorbox[HTML]{E0F4FF}{\textbf{0.60}}
& \colorbox[HTML]{E0F4FF}{\textbf{0.47}}
& \colorbox[HTML]{E0F4FF}{\textbf{0.63}}
& 0.27
& 0.07
& 0.17
& 0.28
& 0.00
& 0.17
& 0.00
& 0.00
& 0.04 \\

XSkill~\cite{xu2023xskillcrossembodimentskill} 
& \colorbox[HTML]{E0F4FF}{\textbf{0.60}}
& 0.13
& 0.43
& 0.39
& 0.53
& 0.07
& 0.00
& 0.07
& 0.17
& 0.03
& 0.00
& 0.00
& 0.10
& 0.03 \\

\textbf{Demo-JEPA (Ours)} 
& 0.33
& 0.17
& 0.43
& 0.31
& 0.60
& \colorbox[HTML]{E0F4FF}{\textbf{0.43}}
& \colorbox[HTML]{E0F4FF}{\textbf{0.37}}
& \colorbox[HTML]{E0F4FF}{\textbf{0.40}}
& \colorbox[HTML]{E0F4FF}{\textbf{0.45}}
& \colorbox[HTML]{E0F4FF}{\textbf{0.20}}
& \colorbox[HTML]{E0F4FF}{\textbf{0.60}}
& \colorbox[HTML]{E0F4FF}{\textbf{0.33}}
& \colorbox[HTML]{E0F4FF}{\textbf{0.30}}
& \colorbox[HTML]{E0F4FF}{\textbf{0.36}} \\
\bottomrule
\end{tabular}}

\end{table}
\begin{table}[!t]
\centering
\vspace{-0.4cm}
\caption{\textbf{Real-world evaluation results.}
We report task success rates on physical robot tasks under the same three-suite evaluation protocol as in simulation.}
\label{tab:real_results}
\resizebox{\linewidth}{!}{
\begin{tabular}{lccccccccc}
\toprule
\multirow{2}{*}[-1.8ex]{\makecell[c]{\textbf{Method}}}
& \multicolumn{3}{c}{\textbf{Behavior Grounding}}
& \multicolumn{3}{c}{\textbf{Cross-Embodiment Bridging}}
& \multicolumn{3}{c}{\textbf{Zero-Shot Generalization}} \\
\cmidrule(lr){2-4} \cmidrule(lr){5-7} \cmidrule(lr){8-10}
& \makecell{Lift\\Cube}
& \makecell{Lift\\Cup}
& \makecell{\textbf{Avg.}}
& \makecell{Remove\\Plate}
& \makecell{Press\\Button}
& \makecell{\textbf{Avg.}}
& \makecell{Remove Pot\\Lid}
& \makecell{Remove\\Pepper}
& \makecell{\textbf{Avg.}} \\
\midrule
VPP~\cite{hu2025videopredictionpolicygeneralist}
& \colorbox[HTML]{E0F4FF}{\textbf{0.55}}
& \colorbox[HTML]{E0F4FF}{\textbf{0.75}}
& \colorbox[HTML]{E0F4FF}{\textbf{0.65}}
& 0.15
& \colorbox[HTML]{E0F4FF}{\textbf{0.90}}
& 0.53
& 0.00
& 0.00
& 0.00 \\

XSkill~\cite{xu2023xskillcrossembodimentskill}
& 0.40
& 0.50
& 0.45
& 0.20
& 0.60
& 0.40
& 0.00
& 0.10
& 0.05 \\

\textbf{Demo-JEPA (Ours)}
& 0.25
& 0.60
& 0.43
& \colorbox[HTML]{E0F4FF}{\textbf{0.30}}
& 0.80
& \colorbox[HTML]{E0F4FF}{\textbf{0.55}}
& \colorbox[HTML]{E0F4FF}{\textbf{0.15}}
& \colorbox[HTML]{E0F4FF}{\textbf{0.35}}
& \colorbox[HTML]{E0F4FF}{\textbf{0.25}} \\

\bottomrule
\end{tabular}}
\vspace{-0.5cm}
\end{table}
\textbf{Simulation Results.}

Table~\ref{tab:eval_results} reports the simulation results across three evaluation suites. 
VPP performs best in behavior grounding, suggesting its advantage in in-domain trajectory learning. 
However, Demo-JEPA shows stronger performance as the distribution shift increases. 
In cross-embodiment bridging, Demo-JEPA achieves an average success rate of 0.45, outperforming VPP and XSkill by 0.17 and 0.28, respectively. 
In zero-shot generalization, Demo-JEPA further achieves 0.36, while VPP and XSkill only reach 0.04 and 0.03. 
These results indicate that the Dreamer Predictor can infer target-compatible latent goals beyond direct action supervision, enabling more robust cross-embodiment transfer under unseen task configurations.

\textbf{Real-World Results.}

In the real-world experiments, VPP also performs best in behavior grounding, where task-specific trajectory regularities are closer to the training distribution. 
Under stronger distribution shifts, Demo-JEPA becomes more effective. 
It achieves the best average success rate of 0.55 in cross-embodiment bridging and reaches 0.25 in zero-shot generalization. 
These findings validate our hypothesis: interpreting cross-embodiment demonstrations as latent goals is more effective than using them as fixed action sequences or skill prototypes. By mapping UR5e prompts to Franka-compatible latent targets, the Dreamer Predictor successfully bridges the embodiment gap.

\subsection{Analysis}
\vspace{-0.2cm}

\begin{table}[!t]
\centering
\caption{
\textbf{Goal reference comparison and architecture ablation in simulation.}
We compare naive, oracle, full Demo-JEPA, and w/o Conv3D variants on simulation tasks.}
\label{tab:oracle_comparison_sim}
\resizebox{\textwidth}{!}{
\begin{tabular}{lcccccccccccccc}
\toprule
\multirow{2}{*}[-1.8ex]{\makecell[c]{\textbf{Method}}}
& \multicolumn{4}{c}{\textbf{Behavior Grounding}}
& \multicolumn{5}{c}{\textbf{Cross-Embodiment Bridging}}
& \multicolumn{5}{c}{\textbf{Zero-Shot Generalization}} \\
\cmidrule(lr){2-5} \cmidrule(lr){6-10} \cmidrule(lr){11-15}
& \makecell{Basketball\\in Hoop}
& \makecell{Change\\Channel}
& \makecell{Close\\Box}
& \makecell{\textbf{Avg.}}
& \makecell{Push\\Button}
& \makecell{Pick\\and Lift}
& \makecell{Slide Block\\to Target}
& \makecell{Pick up\\Cup}
& \makecell{\textbf{Avg.}}
& \makecell{Phone\\on Base}
& \makecell{Close Laptop\\Lid}
& \makecell{Put Rubbish\\in Bin}
& \makecell{Close\\Drawer}
& \makecell{\textbf{Avg.}} \\
\midrule

\textcolor{gray}{V-JEPA 2.1 (Naive)}
& \textcolor{gray}{--}
& \textcolor{gray}{--}
& \textcolor{gray}{--}
& \textcolor{gray}{--}
& \textcolor{gray}{--}
& \textcolor{gray}{--}
& \textcolor{gray}{--}
& \textcolor{gray}{--}
& \textcolor{gray}{--}
& \textcolor{gray}{--}
& \textcolor{gray}{--}
& \textcolor{gray}{--}
& \textcolor{gray}{--}
& \textcolor{gray}{--} \\

\textcolor{gray}{V-JEPA 2.1 (Oracle)}
& \textcolor{gray}{0.43}
& \textcolor{gray}{0.10}
& \textcolor{gray}{0.50}
& \textcolor{gray}{0.34}
& \textcolor{gray}{0.83}
& \textcolor{gray}{0.50}
& \textcolor{gray}{0.27}
& \textcolor{gray}{0.60}
& \textcolor{gray}{0.55}
& \textcolor{gray}{0.33}
& \textcolor{gray}{0.60}
& \textcolor{gray}{0.50}
& \textcolor{gray}{0.23}
& \textcolor{gray}{0.42} \\

\textbf{Demo-JEPA}
& \colorbox[HTML]{E0F4FF}{\textbf{0.33}}
& \colorbox[HTML]{E0F4FF}{\textbf{0.17}}
& \colorbox[HTML]{E0F4FF}{\textbf{0.43}}
& \colorbox[HTML]{E0F4FF}{\textbf{0.31}}
& \colorbox[HTML]{E0F4FF}{\textbf{0.60}}
& \colorbox[HTML]{E0F4FF}{\textbf{0.43}}
& \colorbox[HTML]{E0F4FF}{\textbf{0.37}}
& \colorbox[HTML]{E0F4FF}{\textbf{0.40}}
& \colorbox[HTML]{E0F4FF}{\textbf{0.45}}
& 0.20
& \colorbox[HTML]{E0F4FF}{\textbf{0.60}}
& 0.33
& \colorbox[HTML]{E0F4FF}{\textbf{0.30}}
& \colorbox[HTML]{E0F4FF}{\textbf{0.36}} \\

\text{Demo-JEPA(w/o Conv3D)}
& \colorbox[HTML]{E0F4FF}{\textbf{0.33}}
& 0.03
& 0.27
& 0.21
& \colorbox[HTML]{E0F4FF}{\textbf{0.60}}
& \colorbox[HTML]{E0F4FF}{\textbf{0.43}}
& 0.33
& \colorbox[HTML]{E0F4FF}{\textbf{0.40}}
& 0.44
& \colorbox[HTML]{E0F4FF}{\textbf{0.23}}
& 0.23
& \colorbox[HTML]{E0F4FF}{\textbf{0.40}}
& \colorbox[HTML]{E0F4FF}{\textbf{0.30}}
& 0.29 \\

\bottomrule
\end{tabular}}
\end{table}
\begin{table}[!t]
\centering
\vspace{-0.4cm}
\caption{\textbf{Goal reference comparison and architecture ablation in the real world.}
Following the same protocol as Table~\ref{tab:oracle_comparison_sim}, we evaluate the variants on real-world tasks.}
\label{tab:oracle_comparison_real}
\resizebox{\linewidth}{!}{
\begin{tabular}{lccccccccc}
\toprule
\multirow{2}{*}[-1.8ex]{\makecell[c]{\textbf{Method}}}
& \multicolumn{3}{c}{\textbf{Behavior Grounding}}
& \multicolumn{3}{c}{\textbf{Cross-Embodiment Bridging}}
& \multicolumn{3}{c}{\textbf{Zero-Shot Generalization}} \\
\cmidrule(lr){2-4} \cmidrule(lr){5-7} \cmidrule(lr){8-10}
& \makecell{Lift\\Cube}
& \makecell{Lift\\Cup}
& \makecell{\textbf{Avg.}}
& \makecell{Remove\\Plate}
& \makecell{Press\\Button}
& \makecell{\textbf{Avg.}}
& \makecell{Remove Pot\\Lid}
& \makecell{Remove\\Pepper}
& \makecell{\textbf{Avg.}} \\
\midrule

\textcolor{gray}{V-JEPA 2.1 (Naive)}
& \textcolor{gray}{--}
& \textcolor{gray}{--}
& \textcolor{gray}{--}
& \textcolor{gray}{--}
& \textcolor{gray}{--}
& \textcolor{gray}{--}
& \textcolor{gray}{--}
& \textcolor{gray}{--}
& \textcolor{gray}{--} \\

\textcolor{gray}{V-JEPA 2.1 (Oracle)}
& \textcolor{gray}{0.35}
& \textcolor{gray}{0.75}
& \textcolor{gray}{0.55}
& \textcolor{gray}{0.40}
& \textcolor{gray}{0.75}
& \textcolor{gray}{0.58}
& \textcolor{gray}{0.15}
& \textcolor{gray}{0.40}
& \textcolor{gray}{0.28} \\

\textbf{Demo-JEPA}
& \colorbox[HTML]{E0F4FF}{\textbf{0.25}}
& \colorbox[HTML]{E0F4FF}{\textbf{0.60}}
& \colorbox[HTML]{E0F4FF}{\textbf{0.43}}
& \colorbox[HTML]{E0F4FF}{\textbf{0.30}}
& \colorbox[HTML]{E0F4FF}{\textbf{0.80}}
& \colorbox[HTML]{E0F4FF}{\textbf{0.55}}
& 0.15
& \colorbox[HTML]{E0F4FF}{\textbf{0.35}}
& \colorbox[HTML]{E0F4FF}{\textbf{0.25}} \\

\text{Demo-JEPA(w/o Conv3D)}
& 0.20
& 0.50
& 0.35
& 0.10
& 0.70
& 0.40
& \colorbox[HTML]{E0F4FF}{\textbf{0.20}}
& 0.25
& 0.23 \\

\bottomrule
\end{tabular}}
\vspace{-0.4cm}
\end{table}
\textbf{Goal Reference Comparison with V-JEPA 2.1.}
 We compare Demo-JEPA with two V-JEPA 2.1~\cite{murlabadia2026vjepa21unlockingdense} references that use different planning goals. 
The naive reference directly uses the future latent state from the source demonstration, while the oracle uses target-embodiment ground-truth future trajectories that are unavailable at deployment. 
This comparison tests whether V-JEPA 2.1 latents are directly transferable across embodiments and how closely Demo-JEPA can approach privileged target-future planning.
Tables~\ref{tab:oracle_comparison_sim} and~\ref{tab:oracle_comparison_real} show that the naive reference fails across all tasks, indicating that V-JEPA 2.1 alone does not provide cross-embodiment goal compatibility. 
By contrast, Demo-JEPA closely approaches the oracle using only source demonstrations and current target observations. 
These results show that the Dreamer Predictor converts heterogeneous demonstrations into target-compatible latent goals for planning.

\textbf{Ablations.}
We ablate the Conv3D-based fusion module by replacing it with mean pooling over stacked latent features (Tables~\ref{tab:oracle_comparison_sim} and~\ref{tab:oracle_comparison_real}). 
In simulation, Conv3D's impact correlates with motion complexity. While mean pooling suffices for simpler tasks (\emph{Basketball in Hoop}), Conv3D significantly improves performance on tasks requiring structured, non-trivial motions like twisting (\emph{Change Channel}) or articulated closure (\emph{Close Box}), which defeat permutation-invariant averaging. 
This advantage amplifies in noisy real-world settings with larger embodiment discrepancies; coordination-heavy tasks like \emph{Remove Plate} degrade consistently without Conv3D. Overall, while not strictly required for simple tasks, Conv3D's explicit spatiotemporal modeling is crucial for preserving the structured temporal information needed for complex, articulated, or real-world behaviors.

\begin{table}[!t]
\centering
\caption{\textbf{Imitation learning extension in simulation.}
We report simulation success rates for standard DP, Demo-DP, and the original Demo-JEPA. 
}
\label{tab:dp_extension_sim_results}
\resizebox{\textwidth}{!}{
\begin{tabular}{lcccccccccccccc}
\toprule
\multirow{2}{*}{\textbf{Method}}
& \multicolumn{4}{c}{\textbf{Behavior Grounding}}
& \multicolumn{5}{c}{\textbf{Cross-Embodiment Bridging}}
& \multicolumn{5}{c}{\textbf{Zero-Shot Generalization}} \\
\cmidrule(lr){2-5} \cmidrule(lr){6-10} \cmidrule(lr){11-15}
& \makecell{Basketball\\in Hoop}
& \makecell{Change\\Channel}
& \makecell{Close\\Box}
& \makecell{\textbf{Avg.}}
& \makecell{Push\\Button}
& \makecell{Pick\\and Lift}
& \makecell{Slide Block\\to Target}
& \makecell{Pick up\\Cup}
& \makecell{\textbf{Avg.}}
& \makecell{Phone\\on Base}
& \makecell{Close Laptop\\Lid}
& \makecell{Put Rubbish\\in Bin}
& \makecell{Close\\Drawer}
& \makecell{\textbf{Avg.}} \\
\midrule
DP
& 0.17
& 0.33
& 0.20
& 0.23
& 0.67
& --
& --
& --
& --
& --
& --
& --
& --
& -- \\

Demo-DP
& \colorbox[HTML]{E0F4FF}{\textbf{0.33}}
& \colorbox[HTML]{E0F4FF}{\textbf{0.37}}
& 0.13
& 0.28
& \colorbox[HTML]{E0F4FF}{\textbf{0.83}}
& 0.33
& 0.30
& 0.30
& 0.44
& 0.10
& 0.40
& 0.07
& 0.23
& 0.18 \\

\textbf{Demo-JEPA (Ours)}
& \colorbox[HTML]{E0F4FF}{\textbf{0.33}}
& 0.17
& \colorbox[HTML]{E0F4FF}{\textbf{0.43}}
& \colorbox[HTML]{E0F4FF}{\textbf{0.31}}
& 0.60
& \colorbox[HTML]{E0F4FF}{\textbf{0.43}}
& \colorbox[HTML]{E0F4FF}{\textbf{0.37}}
& \colorbox[HTML]{E0F4FF}{\textbf{0.40}}
& \colorbox[HTML]{E0F4FF}{\textbf{0.45}}
& \colorbox[HTML]{E0F4FF}{\textbf{0.20}}
& \colorbox[HTML]{E0F4FF}{\textbf{0.60}}
& \colorbox[HTML]{E0F4FF}{\textbf{0.33}}
& \colorbox[HTML]{E0F4FF}{\textbf{0.30}}
& \colorbox[HTML]{E0F4FF}{\textbf{0.36}} \\
\bottomrule
\end{tabular}}
\end{table}

\begin{table}[!t]
\centering
\vspace{-0.4cm}
\caption{\textbf{Imitation learning extension in the real world.}
We report real-world success rates for standard DP, Demo-DP, and the original Demo-JEPA. }
\label{tab:dp_extension_real_results}
\resizebox{\textwidth}{!}{
\begin{tabular}{lccccccccc}
\toprule
\multirow{2}{*}{\textbf{Method}}
& \multicolumn{3}{c}{\textbf{Behavior Grounding}}
& \multicolumn{3}{c}{\textbf{Cross-Embodiment Bridging}}
& \multicolumn{3}{c}{\textbf{Zero-Shot Generalization}} \\
\cmidrule(lr){2-4} \cmidrule(lr){5-7} \cmidrule(lr){8-10}
& \makecell{Lift\\Cube}
& \makecell{Lift\\Cup}
& \makecell{\textbf{Avg.}}
& \makecell{Remove\\Plate}
& \makecell{Press\\Button}
& \makecell{\textbf{Avg.}}
& \makecell{Remove Pot\\Lid}
& \makecell{Remove\\Pepper}
& \makecell{\textbf{Avg.}} \\
\midrule
DP
& 0.45
& 0.60
& 0.53
& --
& --
& --
& --
& --
& -- \\

Demo-DP
& \colorbox[HTML]{E0F4FF}{\textbf{0.50}}
& \colorbox[HTML]{E0F4FF}{\textbf{0.80}}
& \colorbox[HTML]{E0F4FF}{\textbf{0.65}}
& \colorbox[HTML]{E0F4FF}{\textbf{0.50}}
& \colorbox[HTML]{E0F4FF}{\textbf{0.95}}
& \colorbox[HTML]{E0F4FF}{\textbf{0.73}}
& 0.10
& 0.20
& 0.15 \\

\textbf{Demo-JEPA (Ours)}
& 0.25
& 0.60
& 0.43
& 0.30
& 0.80
& 0.55
& \colorbox[HTML]{E0F4FF}{\textbf{0.15}}
& \colorbox[HTML]{E0F4FF}{\textbf{0.35}}
& \colorbox[HTML]{E0F4FF}{\textbf{0.25}} \\
\bottomrule
\end{tabular}}
\vspace{-0.4cm}
\end{table}
\textbf{Imitation Learning Extension.}
Finally, we test the Dreamer Predictor's utility in standard imitation learning. In \textbf{Demo-DP}, the predicted target future representation conditions a Diffusion Policy (DP) head~\cite{chi2025diffusion}, explicitly providing demonstration-derived future states instead of relying solely on current observations. 
Tables~\ref{tab:dp_extension_sim_results} and~\ref{tab:dp_extension_real_results} show Demo-DP consistently improves over standard DP in both simulation and real-world settings, proving the predictive latent goals are strong conditioning signals. Furthermore, Demo-DP outperforms VPP and XSkill in zero-shot generalization. 
Comparing Demo-DP with the original Demo-JEPA isolates the impact of the execution module. Demo-DP excels in behavior grounding, as the diffusion policy acts as a strong local action expert under familiar conditions. Conversely, Demo-JEPA is superior in zero-shot generalization, highlighting that while DP is powerful in-domain, planner-based execution remains far more robust under the severe distribution shifts of unseen tasks.

\begin{table}[!t]
\centering
\vspace{-0.2cm}
\caption{\textbf{Scaling study in simulation.}
We report success rates under different data and task scaling ratios. 
(a) summarizes task-level results and suite averages, while (b) visualizes the average trends under data scaling and task scaling.}
\label{tab:scaling_results}

\begin{minipage}[t]{0.58\linewidth}
\vspace{8pt}
\centering
\setlength{\tabcolsep}{3.2pt}
\renewcommand{\arraystretch}{1.05}
\resizebox{\linewidth}{!}{
\begin{tabular}{llcccccccccc}
\toprule
\multirow{2}{*}[-0.6ex]{\makecell[c]{\textbf{Scaling}\\\textbf{Setting}}}
& \multirow{2}{*}[-0.6ex]{\makecell[c]{\textbf{Ratio}}}
& \multicolumn{5}{c}{\textbf{Cross-Embodiment Bridging}}
& \multicolumn{5}{c}{\textbf{Zero-Shot Generalization}} \\
\cmidrule(lr){3-7} \cmidrule(lr){8-12}
& 
& \makecell{Push\\Button}
& \makecell{Pick\\and Lift}
& \makecell{Slide Block\\to Target}
& \makecell{Pick up\\Cup}
& \makecell{\textbf{Avg.}}
& \makecell{Phone\\on Base}
& \makecell{Close Laptop\\Lid}
& \makecell{Put Rubbish\\in Bin}
& \makecell{Close\\Drawer}
& \makecell{\textbf{Avg.}} \\
\midrule
\multirow{2}{*}{Data Scaling}
& 20\%
& 0.50
& 0.23
& 0.00
& 0.33
& 0.27
& 0.10
& 0.37
& \colorbox[HTML]{E0F4FF}{\textbf{0.33}}
& 0.27
& 0.27 \\

& 50\%
& \colorbox[HTML]{E0F4FF}{\textbf{0.60}}
& \colorbox[HTML]{E0F4FF}{\textbf{0.43}}
& 0.20
& 0.27
& 0.38
& 0.13
& 0.40
& \colorbox[HTML]{E0F4FF}{\textbf{0.33}}
& \colorbox[HTML]{E0F4FF}{\textbf{0.30}}
& 0.29 \\

\midrule
\multirow{2}{*}{Task Scaling}
& 20\%
& 0.33
& 0.00
& 0.10
& 0.27
& 0.18
& 0.00
& 0.33
& 0.17
& 0.20
& 0.18 \\

& 50\%
& 0.50
& 0.40
& 0.13
& 0.30
& 0.33
& 0.13
& 0.33
& \colorbox[HTML]{E0F4FF}{\textbf{0.33}}
& 0.20
& 0.25 \\

\midrule
Full
& 100\%
& \colorbox[HTML]{E0F4FF}{\textbf{0.60}}
& \colorbox[HTML]{E0F4FF}{\textbf{0.43}}
& \colorbox[HTML]{E0F4FF}{\textbf{0.37}}
& \colorbox[HTML]{E0F4FF}{\textbf{0.40}}
& \colorbox[HTML]{E0F4FF}{\textbf{0.45}}
& \colorbox[HTML]{E0F4FF}{\textbf{0.20}}
& \colorbox[HTML]{E0F4FF}{\textbf{0.60}}
& \colorbox[HTML]{E0F4FF}{\textbf{0.33}}
& \colorbox[HTML]{E0F4FF}{\textbf{0.30}}
& \colorbox[HTML]{E0F4FF}{\textbf{0.36}} \\
\bottomrule
\end{tabular}}
\par\vspace{9pt}
{\footnotesize\textbf{(a)}}
\end{minipage}
\hfill
\begin{minipage}[t]{0.4\linewidth}
\vspace{0pt}
\centering
\includegraphics[width=\linewidth]{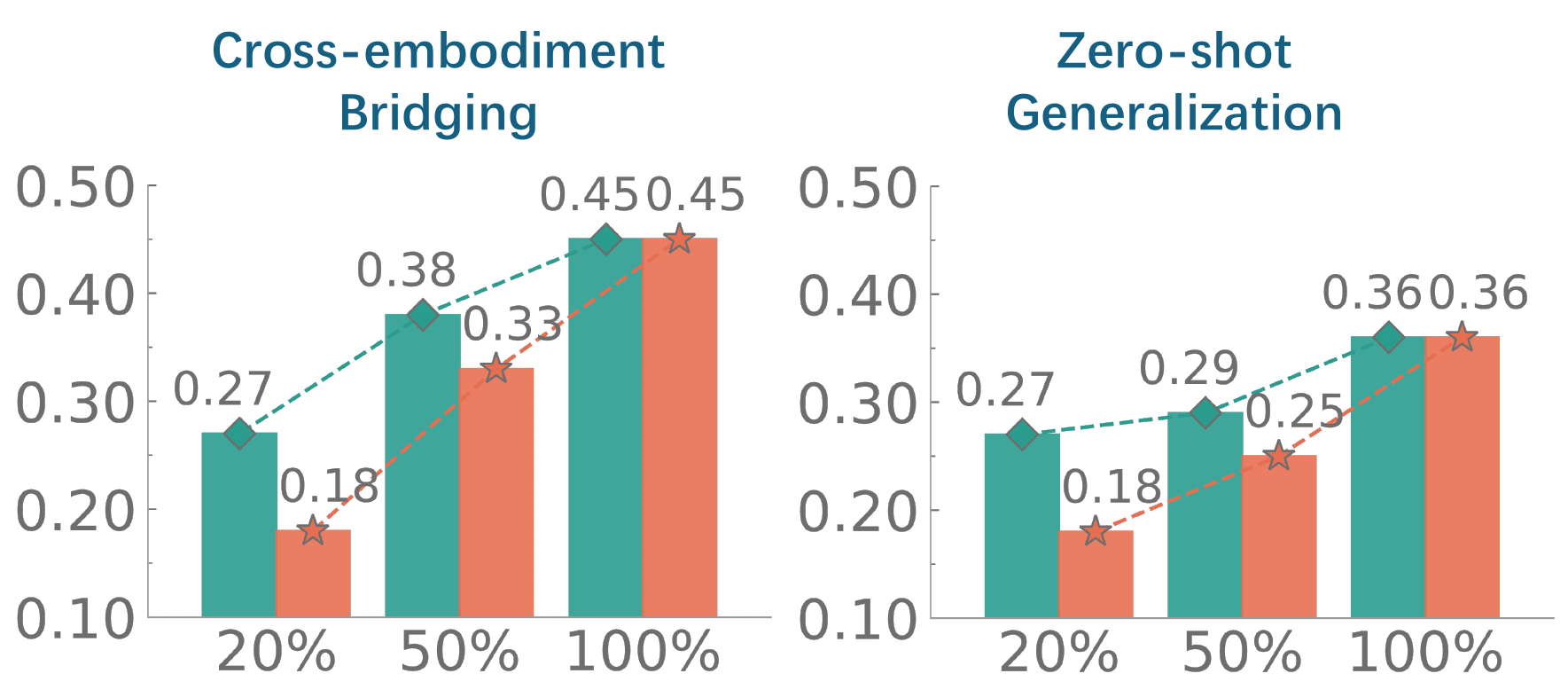}
{\footnotesize\textbf{(b)}}
\end{minipage}
\vspace{-0.55cm}
\end{table}

\textbf{Scaling Study.}
Table~\ref{tab:scaling_results} investigates how the scale of paired visual trajectories affects the Dreamer Predictor. We isolate two scaling dimensions: data scaling, which reduces the number of paired episodes per task to 20\% or 50\%, and task scaling, which reduces the number of task categories to 20\% or 50\%. 
The results show that task diversity is more important than per-task episode volume. At the 20\% scale, data scaling outperforms task scaling in both cross-embodiment bridging (0.27 vs. 0.18) and zero-shot generalization (0.27 vs. 0.18). The same trend holds at the 50\% scale, where data scaling achieves higher average success rates than task scaling in cross-embodiment bridging (0.38 vs. 0.33) and zero-shot generalization (0.29 vs. 0.25). These results indicate that exposure to diverse task semantics is the primary driver of transferable source-to-target latent mappings, while increasing intra-task trajectories further improves predictor robustness.
\section{Conclusion and Limitation}\label{sec:con}
\vspace{-0.2cm}

We present \textbf{Demo-JEPA}, a framework reframing cross-embodiment imitation as \textbf{latent goal planning}. By decoupling task intent from execution via a JEPA-based world model, our method achieves robust zero-shot generalization across diverse morphologies. However, performance is constrained by modeling bottlenecks of current action-conditioned world models, limiting efficacy in complex, high-precision tasks. Additionally, the pipeline still requires temporal or progress-aware alignment during training. Future work will focus on enhancing world model fidelity and developing fully unaligned learning strategies.
{
\bibliographystyle{unsrt}
\bibliography{reference}
}
\appendix
\clearpage
\appendix
\section{Preliminary}

 We briefly review the key building blocks of our approach: action-conditioned world models for modeling robot and environment dynamics, and the cross-entropy method (CEM) for trajectory optimization.

\subsection{Action-Conditioned World Models}\label{app:wm}

Action-conditioned world models learn predictive latent dynamics conditioned on robot actions. 
Given a trajectory dataset
\[
\mathcal{D} = \{\tau_k\}_{k=1}^{N},
\]
where each trajectory is defined as
\[
\tau = \{(o_t, s_t, a_t)\}_{t=1}^{T},
\]
$o_t$ denotes the visual observation, $s_t$ denotes the robot state (e.g., proprioceptive state), and $a_t$ denotes the action executed at time step $t$.

An encoder $E(\cdot)$ first maps observations into latent representations:
\[
z_t = E(o_t).
\]

A dynamics predictor $F_{wm}(\cdot)$ then models future latent transitions conditioned on the current latent state, robot state, and action:
\[
\hat{z}_{t+1} = F_{wm}(z_t, s_t, a_t).
\]

To improve long-horizon dynamics modeling, the predictor can be recursively rolled out using previously predicted latent states:
\[
\hat{z}_{t+k} = F_{wm}(\hat{z}_{t+k-1}, s_{t+k-1}, a_{t+k-1}),
\quad k > 1.
\]
\subsection{Cross entropy method}\label{app:cem}

 Given a learned action-conditioned world model, we aim to infer an action sequence that drives the predicted future latent state toward a target latent representation $z_{gt}$. 
 Formally, we seek an action sequence
 \[
 \mathbf{a}_{t:t+H-1}^*
 \]
 that minimizes the latent prediction error:
 \[
 \mathbf{a}_{t:t+H-1}^*
 =
 \arg\min_{\mathbf{a}_{t:t+H-1}}
 d\left(
 F_{wm}(z_t, s_t, \mathbf{a}_{t:t+H-1}),
 z_{gt}
 \right),
 \]
 where $d(\cdot,\cdot)$ denotes a latent distance metric such as the $\ell_1$ distance.

 Since directly inverting the nonlinear dynamics predictor $F_{wm}(\cdot)$ is intractable, we employ the Cross-Entropy Method (CEM) to iteratively optimize action sequences.

 Specifically, CEM maintains a Gaussian sampling distribution parameterized by mean $M$ and standard deviation $S$. 
At each iteration, we sample $N$ candidate action sequences:
 \[
 \mathbf{a}_{1:N}
 \sim
 \mathcal{N}(M, S).
 \]

 Each candidate action sequence is rolled out through the world model to obtain predicted future latent states:
 \[
 \hat{z}_{1:N}
 =
 F_{wm}(z_t, s_t, \mathbf{a}_{1:N}).
 \]

 We then evaluate each candidate using the latent prediction objective:
 \[
 \mathcal{L}_{plan}
 =
 d(\hat{z}, z_{gt}).
 \]

 The top-$K$ candidates with the lowest planning loss are selected as elites to update the sampling distribution:
 \[
 M \leftarrow
 \beta M + (1-\beta) M_{\text{elite}},
 \]
 \[
 S \leftarrow
 \beta S + (1-\beta) S_{\text{elite}},
 \]
 where $\beta$ denotes the momentum coefficient.

 After multiple iterations, the optimized mean action sequence is used as the final planned action. Although iterative planning with CEM introduces additional computational cost, it avoids directly projecting high-level representations into actions. Instead, actions are optimized online through latent dynamics rollout, which provides stronger adaptability under environment and embodiment shifts.

\section{Pseudocode}
We present detailed pseudocode in this appendix to clearly and systematically describe the core algorithms of our framework, including the Cross-Entropy Method (CEM) optimization, as well as the training and inference procedures. The aim is to provide an unambiguous and implementation-oriented specification that complements the high-level descriptions in the main text.
\begin{algorithm}[H]
\caption{Cross-Entropy Method (CEM) for Latent Trajectory Optimization}
\label{alg:cem}
\linespread{1.2}\selectfont
\begin{algorithmic}[1]
\Require 
    Current latent state $z_t$; 
    Current robot state $s_t$; 
    Target latent $z_{gt}$; 
    Dynamics predictor $F_{wm}(\cdot)$; 
    Planning horizon $H$; 
    Iterations $L$; 
    Population size $N$; 
    Number of elites $K$; 
    Momentum $\beta$.

\Ensure 
    Optimized action sequence $\mathbf{a}_{t:t+H-1}^*$.

\State \textbf{Initialize} mean $M \in \mathbb{R}^{H \times \dim(a)}$ and standard deviation $S \in \mathbb{R}^{H \times \dim(a)}$.
\State \textit{// e.g., $M \leftarrow \mathbf{0}$, $S \leftarrow \mathbf{I}$}

\For{$iter = 1$ \textbf{to} $L$}
    \State \textbf{Sample} $N$ candidate action sequences:
    \State $\mathcal{A} = \{\mathbf{a}_{1}, \dots, \mathbf{a}_{N}\}$, where $\mathbf{a}_i \sim \mathcal{N}(M, \operatorname{diag}(S^2))$
    
    \For{each candidate $\mathbf{a}_i \in \mathcal{A}$}
        \State \textbf{Rollout} the world model to predict future latent state:
        \State $\hat{z}_{t+H}^{(i)} = F_{wm}(z_t, s_t, \mathbf{a}_i)$ \Comment{Recursive prediction using Equation~\ref{app:wm}}
        \State \textbf{Evaluate} planning loss:
        \State $\mathcal{L}_i = d(\hat{z}_{t+H}^{(i)}, z_{gt})$ \Comment{Using $\ell_1$ or $\ell_2$ distance}
    \EndFor
    
    \State \textbf{Identify} indices of $K$ sequences with lowest loss (elites):
    \State $\mathcal{E} = \operatorname{arg-top-}K \text{ smallest } \{\mathcal{L}_1, \dots, \mathcal{L}_N\}$
    
    \State \textbf{Compute} statistics of elite sequences:
    \State $M_{\text{elite}} = \frac{1}{K} \sum_{i \in \mathcal{E}} \mathbf{a}_i$
    \State $S_{\text{elite}} = \sqrt{\frac{1}{K} \sum_{i \in \mathcal{E}} (\mathbf{a}_i - M_{\text{elite}})^2}$
    
    \State \textbf{Update} sampling distribution with momentum:
    \State $M \leftarrow \beta M + (1 - \beta) M_{\text{elite}}$
    \State $S \leftarrow \beta S + (1 - \beta) S_{\text{elite}}$
\EndFor

\State \Return $\mathbf{a}^* = M$
\end{algorithmic}
\end{algorithm}
\begin{algorithm}[tbp]
\caption{Two-Stage Training of Demo-JEPA}
\label{alg:train}
\begin{algorithmic}[1]
\Require 
    Source dataset $\mathcal{D}^s$, Target dataset $\mathcal{D}^t$; 
    Pretrained encoder $E$, world model $F_{wm}$; 
    Perturbation radius $r$, epochs $E_1$, $E_2$.
\Ensure 
    Trained Dreamer Predictor $\mathcal{P}_\theta$, fine-tuned $F_{wm}$.

\State \textbf{Stage 1: Train Dreamer Predictor}
\For{$epoch = 1$ to $E_1$}
    \State Sample target $o^t_k$, $o^t_{k+n}$ and source $o^s_k$, $o^s_{k+n}$
    \State Sample perturbation $\delta \sim \mathcal{U}(-r,r)$, set $i = k+\delta$, $j = k+n$
    \State Encode: $z^t_k = E(o^t_k)$, $z^s_i = E(o^s_i)$, $z^s_j = E(o^s_j)$
    \State Compute embodiment cross-attention: $f_{\text{emb}} = \mathrm{Attn}(z^t_k, z^s_i, z^s_i)$
    \State Compute motion cross-attention: $f_{\text{mot}} = \mathrm{Attn}(z^s_j, z^s_i, z^s_i)$
    \State Fuse: $f_{\text{fused}} = \phi([z^t_k \oplus f_{\text{emb}} \oplus f_{\text{mot}}])$
    \State Predict latent goal: $\hat z_{\text{goal}} = \mathcal{P}_\theta(f_{\text{fused}})$
    \State Compute loss: $\mathcal{L}_{\text{pred}} = \|\hat z_{\text{goal}} - E(o^t_{k+n})\|_2^2$
    \State Update $\mathcal{P}_\theta$ to minimize $\mathcal{L}_{\text{pred}}$
\EndFor

\State \textbf{Stage 2: Action Co-Training}
\State Freeze $\mathcal{P}_\theta$
\For{$epoch = 1$ to $E_2$}
    \State Compute $\hat z_{\text{goal}}$ using frozen $\mathcal{P}_\theta$ (same as Stage 1)
    \State Sample action sequence $\mathbf{a} = a_{k:k+n-1}^t$
    \State $\tilde z = F_{wm}(z^t_k, s^t_k, \mathbf{a})$
    \State Compute planning loss: $\mathcal{L}_{\text{plan}} = \|\tilde z - \hat z_{\text{goal}}\|_2^2$
    \State Update $F_{wm}$ to minimize $\mathcal{L}_{\text{plan}}$
\EndFor

\State \Return $\mathcal{P}_\theta$, $F_{wm}$
\end{algorithmic}
\end{algorithm}
\begin{algorithm}[tbp]
\caption{Inference with Adaptive Goal Updating}
\label{alg:inference}
\begin{algorithmic}[1]
\Require 
    Source demonstration $\tau^s = \{o^s_1,\dots,o^s_N\}$; 
    Initial target observation $o^t_0$; 
    Encoder $E$; Dreamer Predictor $\mathcal{P}_\theta$; 
    World model $F_{wm}$; CEM planner (Algorithm~\ref{alg:cem}); 
    Temporal offset $\Delta$; threshold $\epsilon$.
\Ensure 
    Executed action sequence for task completion.

\State $i \gets 1$
\State $z^t \gets E(o^t_0)$
\State Encode source pair: $z^s_{\text{cur}} \gets E(o^s_i)$, $z^s_{\text{fut}} \gets E(o^s_{i+\Delta})$
\State Compute initial latent goal:
\State \quad $f_{\text{emb}} = \mathrm{Attn}(z^t, z^s_{\text{cur}}, z^s_{\text{cur}})$
\State \quad $f_{\text{mot}} = \mathrm{Attn}(z^s_{\text{fut}}, z^s_{\text{cur}}, z^s_{\text{cur}})$
\State \quad $f_{\text{fused}} = \phi([z^t \oplus f_{\text{emb}} \oplus f_{\text{mot}}])$
\State \quad $\hat z_{\text{goal}} = \mathcal{P}_\theta(f_{\text{fused}})$

\While{$i + \Delta \le N$}
    \State Optimize action sequence via CEM:
    \State \quad $\mathbf{a}^* \gets \mathrm{CEM}(F_{wm}, z^t, s^t, \hat z_{\text{goal}})$
    \State Execute first action $a^*_0$, observe new $o^t_{\text{next}}$, $s^t_{\text{next}}$
    \State $z^t_{\text{next}} \gets E(o^t_{\text{next}})$
    \State Compute discrepancy $D = d(z^t_{\text{next}}, \hat z_{\text{goal}})$
    \If{$D < \epsilon$}
        \State $i \gets i + 1$
        \If{$i + \Delta \le N$}
            \State Update source pair: $z^s_{\text{cur}} \gets E(o^s_i)$, $z^s_{\text{fut}} \gets E(o^s_{i+\Delta})$
            \State Recompute $\hat z_{\text{goal}}$ (same as initialization)
        \EndIf
    \EndIf
    \State $z^t \gets z^t_{\text{next}}$, $s^t \gets s^t_{\text{next}}$
\EndWhile

\State \Return success
\end{algorithmic}
\end{algorithm}
\FloatBarrier

\section{Training Details}\label{app:training_details}

We present the training details of Demo-JEPA and Demo-DP. To improve clarity, we separate shared configurations from stage-specific settings.

\subsection{Shared Training Configuration}

All stages use identical data preprocessing, augmentation, and optimization settings unless otherwise specified.

\begin{table}[htbp]
\centering
\caption{Shared training configuration across all stages.}
\label{tab:shared_config}
\begin{tabular}{p{0.28\textwidth} p{0.20\textwidth} p{0.45\textwidth}}
\toprule
\textbf{Component} & \textbf{Value} & \textbf{Description} \\
\midrule
Crop size & 256 & Input image resolution \\
Patch size & 16 & ViT patch size \\
Tubelet size & 2 & Temporal grouping of frames \\
FPS & 5 & Video sampling rate \\
Context frames & 8 & Sequence length per sample \\
\midrule
Optimizer & AdamW & Default optimizer \\
Base learning rate & 4.25e-4 & Peak learning rate \\
Warmup epochs & 15 & Linear warmup schedule \\
Total epochs & 315 & Training duration \\
Weight decay & 0.04 & L2 regularization \\
\midrule
Augmentation & Disabled & No stochastic augmentation \\
\bottomrule
\end{tabular}
\end{table}

\FloatBarrier

\subsection{Demo-JEPA}

\subsubsection{Stage 0: Action-conditioned world model}

We follow the training setup of VJEPA2 with minor modifications.

\begin{table}[htbp]
\centering
\caption{Stage-specific parameters for AC world model.}
\label{tab:jepa_stage0}
\begin{tabular}{p{0.28\textwidth} p{0.20\textwidth} p{0.45\textwidth}}
\toprule
\textbf{Parameter} & \textbf{Value} & \textbf{Description} \\
\midrule
Batch size & 24 & Per-GPU batch size \\
Predictor depth & 24 & Transformer layers \\
Embed dim & 1024 & Hidden dimension \\
Attention heads & 16 & Multi-head attention \\
Frame-causal & True & Temporal causal masking \\
Modality embedding & True & Image/video separation \\
\bottomrule
\end{tabular}
\end{table}

\FloatBarrier

\subsubsection{Stage 1: Latent goal predictor training}

\begin{table}[htbp]
\centering
\caption{Stage-specific parameters for latent goal predictor.}
\label{tab:jepa_stage1}
\begin{tabular}{p{0.28\textwidth} p{0.20\textwidth} p{0.45\textwidth}}
\toprule
\textbf{Parameter} & \textbf{Value} & \textbf{Description} \\
\midrule
Batch size & 16 & Per-GPU batch size \\
Fusion type & Conv3D & Feature fusion mechanism \\
Num self-attn blocks & 4 & Transformer refinement layers \\
MLP ratio & 4.0 & Expansion ratio \\
Up dim & 64 & Projection dimension \\
Norm layer & LayerNorm & Normalization type \\
Init std & 0.02 & Weight initialization scale \\
\bottomrule
\end{tabular}
\end{table}

\FloatBarrier

\subsubsection{Stage 2: Action co-training}

\begin{table}[htbp]
\centering
\caption{Stage-specific parameters for action co-training.}
\label{tab:jepa_stage2}
\begin{tabular}{p{0.28\textwidth} p{0.20\textwidth} p{0.45\textwidth}}
\toprule
\textbf{Parameter} & \textbf{Value} & \textbf{Description} \\
\midrule
Batch size & 16 & Per-GPU batch size \\
Predictor depth & 24 & Transformer layers \\
Embed dim & 1024 & Hidden dimension \\
Attention heads & 16 & Multi-head attention \\
Frame-causal & True & Temporal causal masking \\
Modality embedding & True & Image/video separation \\
\bottomrule
\end{tabular}
\end{table}

\FloatBarrier

\subsection{Demo-DP}

\subsubsection{Action head fine-tuning}

\begin{table}[htbp]
\centering
\caption{Diffusion head configuration for Demo-DP.}
\label{tab:dp_stage}
\begin{tabular}{p{0.28\textwidth} p{0.20\textwidth} p{0.45\textwidth}}
\toprule
\textbf{Parameter} & \textbf{Value} & \textbf{Description} \\
\midrule
Batch size & 16 & Per-GPU batch size \\
Condition dim & 1408 & Encoder output dimension \\
Horizon & 1 & Action prediction steps \\
Condition steps & 256 & Context tokens \\
Diffusion steps & 100 & Noise schedule steps \\
Beta schedule & squaredcos\_cap\_v2 & Noise schedule type \\
Prediction type & epsilon & Noise prediction target \\
\midrule
Layers & 12 & Transformer depth \\
Condition layers & 8 & Conditioning blocks \\
Heads & 4 & Attention heads \\
Embed dim & 384 & Hidden dimension \\
Dropout (attn) & 0.3 & Attention dropout \\
Causal attention & True & Autoregressive masking \\
Time conditioning & True & Timestep embedding \\
\bottomrule
\end{tabular}
\end{table}

\FloatBarrier

\subsection{Compute Budget}

We summarize the compute budget in Table~\ref{tab:compute_budget}.

\begin{table}[htbp]
\centering
\caption{Compute budget of Demo-JEPA and Demo-DP training stages.}
\label{tab:compute_budget}
\begin{tabular}{l c c}
\toprule
\textbf{Stage} & \textbf{GPU} & \textbf{Training Time} \\
\midrule
\multicolumn{3}{c}{\textbf{Demo-JEPA}} \\
\midrule
Stage 0: World model & 8$\times$A100 & 7 days \\
Stage 1: Goal predictor & 8$\times$A100 & 2.5 days \\
Stage 2: Co-training & 8$\times$A100 & 1 day \\
\midrule
\multicolumn{3}{c}{\textbf{Demo-DP}} \\
\midrule
Action head fine-tuning & 8$\times$A100 & 1 day \\
\bottomrule
\end{tabular}
\end{table}

\FloatBarrier
\section{Real World Setup} \label{app:real_world_setup}
\begin{figure*}[t]
    \centering
    \includegraphics[width=\textwidth]{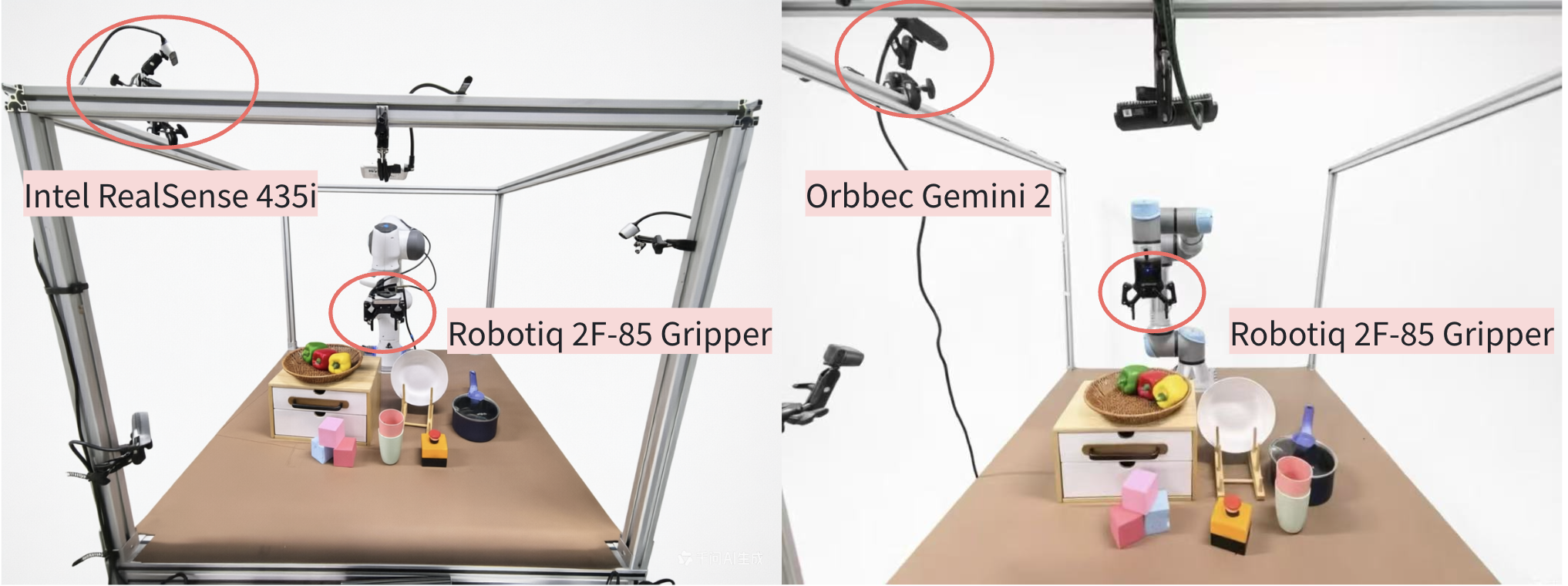}
    \caption{
    \textbf{Real world experiment environment setup.} Franka workspace(left) and UR5e workspace(right).
    }
    \label{fig:real_world_setup}
\end{figure*}

The physical deployment of Demo-JEPA is designed to evaluate cross-embodiment imitation under a minimal and consistent perception setting across heterogeneous robotic platforms.

As shown in Figure~\ref{fig:real_world_setup}, we consider two embodiments: a Franka emika manipulator as the target embodiment and a UR arm as the source embodiment. Both systems are equipped with a Robotiq 2F-85 parallel gripper.

For perception, we use a single fixed third-person RGB camera mounted at a right-shoulder viewpoint relative to the workspace. Specifically, the Franka setup uses an Intel RealSense D435i, while the UR setup uses an Orbbec Gemini 2. Notably, the two cameras are not explicitly calibrated across embodiments; instead, they are roughly placed to ensure that the field of view covers both the robot manipulator and the operation plane, without enforcing precise geometric alignment. There are no wrist-mounted or additional ego-centric cameras are used; both FR3 and UR policies operate solely under third-person visual input.
\section{Broader Impact} \label{app:impacts}

This work aims to improve cross-embodiment robotic imitation learning by enabling transferable goal understanding across heterogeneous agents. Such capabilities may reduce the cost of robot adaptation and improve the scalability of embodied AI systems in domains such as automation, assistance, and training.

At the same time, increasingly capable robotic systems may introduce risks related to unintended behavior transfer, deployment reliability, and safety in real-world environments. Our work is evaluated only in controlled simulation and laboratory settings, and additional safeguards and validation would be necessary before deployment in safety-critical applications.


\end{document}